\documentclass[10pt,twocolumn,letterpaper]{article}

\usepackage{cvpr}              
\definecolor{cvprblue}{rgb}{0.21,0.49,0.74}
\usepackage[pagebackref,breaklinks,colorlinks,allcolors=cvprblue]{hyperref}

\def\confName{CVPR}
\def\confYear{2027}

\title{RoboDuo: Attempt, Check and Atomic Recovery through Code for Vision-Language-Action Models}

\usepackage{marvosym}
\newcommand{\corr}{\mbox{\Letter}}

\author{Jie Wang\textsuperscript{1,2},~
Bo Zhang\textsuperscript{1,\corr},~
Renzhi Lu\textsuperscript{2},~
Jianyu Chen\textsuperscript{1,3}\\[0.4mm]
\textsuperscript{1}Robotera\quad
\textsuperscript{2}Huazhong University of Science and Technology\quad
\textsuperscript{3}Tsinghua University\\[0.2mm]
{\normalsize bo.zhang@robotera.com}
}

\usepackage{algorithm}
\usepackage{algorithmic}
\usepackage{booktabs}
\usepackage{tabularx}

\usepackage[T1]{fontenc}
\usepackage{listings}
\usepackage{xcolor}

\lstdefinestyle{roboduo}{
  language=Python,
  basicstyle=\fontfamily{ptm}\selectfont\footnotesize,
  columns=fullflexible,
  keepspaces=true,
  keywordstyle=\bfseries,
  morekeywords={not,and},
  emph={vla_execute_fulltask,vla_execute_atomic,vla_execute,is_task_complete},
  emphstyle=\bfseries,
  commentstyle=\color{black!55}\itshape,
  stringstyle=\color{black},
  literate={"}{\textquotedbl}1 {_}{\textunderscore}1,
  backgroundcolor=\color{black!5},
  frame=single,
  rulecolor=\color{black!35},
  framesep=4pt,
  xleftmargin=4pt,
  xrightmargin=4pt,
  breaklines=true,
  breakindent=1.5em,
  showstringspaces=false,
  aboveskip=0pt,
  belowskip=0pt,
}

\begin{document}
\maketitle
\begin{abstract}

Vision-Language-Action (VLA) models map a language instruction and the current observation directly to actions, without checking whether the instruction has been completed.
For long-horizon tasks, a single failed step therefore goes unnoticed and leaves the whole task unfinished.
Such failed episodes, however, often end with the task partially complete. Moreover, the unfinished steps are reusable atomic skills that recur across demonstrations. A failed attempt can thus be recovered by composing with atomic skills.
We propose RoboDuo, a closed-loop dual-system framework that uses code as the interface between a coding agent and two VLAs.
The agent writes a program that lets a full-task VLA attempt the whole task and checks whether it is complete; on failure, an atomic VLA finishes the remaining steps.
Since the agent emits no actions and only decides which VLA for execution, RoboDuo wraps an existing full-task VLA without updating its weights.
To train the atomic VLA and the coding agent with the completion-checking ability, we build LIBERO-Atomic, LIBERO-Code, and LIBERO-Judge from a single set of LIBERO demonstrations.
RoboDuo consistently improves X-VLA and $\pi_{0.5}$ on LIBERO and LIBERO-PRO, with the largest gains on long-horizon tasks, and recovers failed rollouts of a frozen full-task VLA on a real robot.

\end{abstract}    
\section{Introduction}
\label{sec:intro}

\begin{figure*}[t]
    \centering
    \includegraphics[width=1\linewidth]{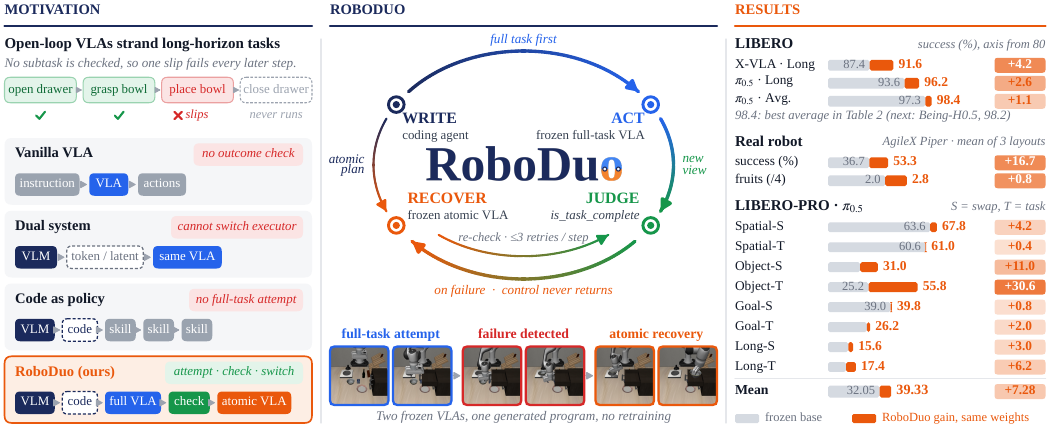}
\caption{\textbf{Overview of RoboDuo and its results.}
Left: open-loop VLAs fail long-horizon tasks after one unnoticed slip; token/latent dual systems cannot switch executor, and code-as-policy skips the full-task attempt.
Center: RoboDuo uses a coding agent to \emph{Write} a program, a frozen full-task VLA to \emph{Act}, \texttt{is\_task\_complete} to \emph{Judge}, and a frozen atomic VLA to \emph{Recover} with at most three retries per step; control does not return.
Right: matched gains on LIBERO, LIBERO-PRO (S/T), and a real robot.
Bottom: one rollout from full-task attempt through failure detection to atomic recovery.}
    \label{fig:teaser}
\end{figure*}

\begin{figure*}[t]
    \centering
    \includegraphics[width=1\linewidth]{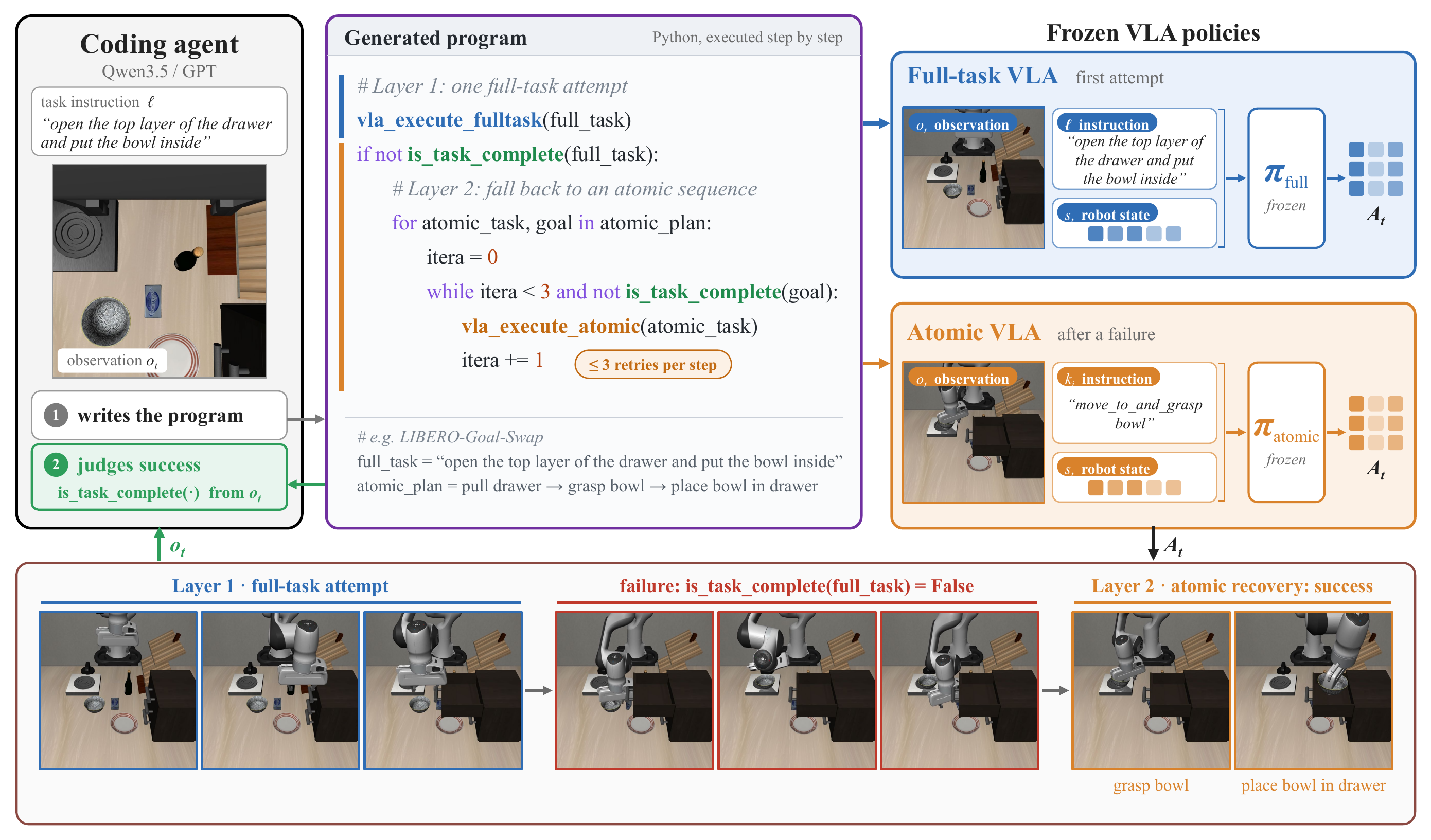}
\caption{\textbf{RoboDuo architecture: a generated program dispatches two VLA policies.}
Given the instruction and the current observation, the coding agent writes a two-layer program.
Layer~1 makes one full-task attempt with $\pi_{\mathrm{full}}$; if \texttt{is\_task\_complete} returns false, Layer~2 executes an atomic plan with $\pi_{\mathrm{atomic}}$, retrying each step at most three times until its goal is verified.
Both policies stay frozen, so the same program wraps X-VLA and $\pi_{0.5}$ without retraining.
The bottom row shows one LIBERO-Goal-Swap rollout: after the full-task attempt fails, the atomic VLA grasps the bowl and places it in the drawer.}
    \label{fig:method}
\end{figure*}


Vision-Language-Action (VLA) models~\cite{brohan2023rt2,kim2024openvla,black2024pi0,intelligence2025pi05,zheng2025xvla,zhao2023act} map a language instruction and the current observation directly to low-level actions, and have become a standard approach to language-conditioned manipulation. The policy is re-queried on each new observation, yet nothing checks whether the current observation satisfies the intended goal. Execution therefore continues under the same instruction even after a step has failed. Consider an object that slips from the gripper during transport. The post-slip state is rare in the demonstrations, so the policy rarely re-grasps and may instead continue its learned trajectory, for example by placing with an empty gripper. On a long-horizon task, that unnoticed failure leaves the remaining steps unfinished, and the episode fails even though earlier progress was correct.

One route is to place this decision-making functionality inside the action model.
Systems built on GPT-6-Astra~\cite{openai2026astra} use one large model to generate numerical robot actions, judge the outcome during execution, and emit a correction within the same forward pass~\cite{galbot2026astra}.
However, the scale that enables such judgment and correction also slows every inference step.
Even steps that a fast policy could execute unchanged must wait on the large model, and the resulting latency falls short of the control rate required for physical deployment.

Dual-system architectures instead place the decision model outside the action model.
In Helix and related systems, a high-level model passes a language command, a token, or a latent vector to a single low-level policy~\cite{figure2025helix,intelligence2025pi05,shi2025hirobot,bjorck2025gr00t,bu2024robodual,zhang2024hirt,cui2025openhelix,shentu2024lcb}.
This interface changes what the low-level policy is conditioned on, but it cannot express a failed attempt, a retry, or a switch to a different executor.
Code-as-policy methods provide exactly this control flow, in which an agent writes a program that tests the outcome of each skill and retries it on failure~\cite{liang2023cap,huang2023voxposer,singh2023progprompt,huang2024rekep,liang2026robofoundry}.
Because these programs call primitives or VLA skills from the first step, they largely discard the long-horizon competence of a full-task VLA.
Code is therefore better used to wrap one attempt by that VLA, and to close the loop only after the attempt fails.

Building on this principle, we propose RoboDuo, a dual-system framework in which a coding agent uses code as the interface to two frozen VLAs (Fig.~\ref{fig:teaser}).
The agent serves as an orchestrator and writes a task program.
The program first assigns the task to the full-task VLA for one attempt, after which \texttt{is\_task\_complete} evaluates the new observation.
If the task fails, the agent plans the unfinished steps from the current observation as atomic instructions.
An atomic VLA then executes these instructions to the end of the task, with at most three attempts per step, and the control does not return to the full-task VLA.
We employ a separate atomic VLA since each atomic instruction describes a short segment already present in the demonstrations, whereas the full-task VLA is trained only on full-task instructions.
We construct LIBERO-Atomic, LIBERO-Code, and LIBERO-Judge from the same demonstrations to train the atomic VLA, the coding agent, and the completion check, respectively.
We instantiate RoboDuo on X-VLA and $\pi_{0.5}$ without updating the weights of either full-task VLA.

On LIBERO-Long, RoboDuo raises the success rate of X-VLA from 87.4\% to 91.6\% and that of $\pi_{0.5}$ from 93.6\% to 96.2\%.
On LIBERO-PRO, RoboDuo raises the mean success of $\pi_{0.5}$ from 32.05\% to 39.33\%, with the largest gain on Object-T ($25.2\rightarrow 55.8$).
On a real robot, RoboDuo raises success from 36.7\% to 53.3\% over three fruit-arrangement layouts, without updating the frozen full-task VLA.


Our main contributions are summarized as follows:
\begin{itemize}
    \item We propose RoboDuo, a closed-loop dual system that uses code, rather than a token or a latent vector, as its interface. When the full-task VLA fails, an atomic VLA executes the remaining steps.
    \item We construct LIBERO-Atomic, LIBERO-Code, and LIBERO-Judge from the same successful demonstrations to train the atomic VLA, the coding agent, and the completion check, respectively.
    \item We instantiate RoboDuo on X-VLA and $\pi_{0.5}$ and evaluate it on LIBERO, LIBERO-PRO, and a real robot. The gains concentrate on long-horizon tasks that fail because a single subtask fails.
\end{itemize}
\section{Related Work}
\label{sec:related}

\textbf{Vision-Language-Action Models.}
VLA models learn end-to-end mappings from language instructions and visual observations to low-level actions, and existing approaches differ mainly in how they represent actions.
Autoregressive models discretize actions into tokens and predict them sequentially with a vision-language backbone~\cite{brohan2023rt1,brohan2023rt2,kim2024openvla}.
Continuous-control models instead attach an action expert or decoder that regresses or denoises future actions~\cite{zhao2023act,octo2024,black2024pi0,intelligence2025pi05,zheng2025xvla}.
Subsequent work inserts intermediate reasoning before action generation, including textual chain-of-thought, visual subgoals, motion foresight, and retrieved trajectory memory~\cite{zawalski2024ecot,zhao2025cotvla,lin2026hifvla,li2026optimusvla}.
These methods improve the quality of a single attempt within the action model.
RoboDuo is complementary to them: it uses such a model, without modification, as the frozen full-task VLA.

\textbf{Dual Systems and Code as Policies.}
Dual-system architectures place a slow vision-language model above a fast policy.
In Helix and related systems, the upper model passes a language command, a token, or a latent vector to that policy~\cite{figure2025helix,intelligence2025pi05,shi2025hirobot,bjorck2025gr00t,zhang2024hirt,bu2024robodual,shentu2024lcb,cui2025openhelix}.
This interface changes only the conditioning of the same executor, and these systems typically lack an explicit completion check.
By contrast, systems built on GPT-6-Astra let one large model emit the actions directly, so every control step incurs the inference latency of that model~\cite{galbot2026astra}.
Code-as-policy methods write programs over control primitives, situated plans, value maps, or keypoint constraints~\cite{liang2023cap,singh2023progprompt,huang2023voxposer,huang2024rekep}.
More recent agents also call frozen VLA policies as skills~\cite{liang2026robofoundry}.
In these systems, the program serves as the plan from the first step, so no full-task policy ever attempts the task as a whole.
RoboDuo also uses code as the interface, but its coding agent emits no actions.
The full-task VLA first attempts the whole task, and the atomic VLA takes over only after the completion check fails.

\textbf{Failure Correction and Recovery.}
Several methods intervene around a failure and then let the original policy continue.
FPC-VLA corrects the direction and magnitude of the current action when the gripper state changes~\cite{yang2025fpcvla}.
Spotter repairs a confirmed error with motion primitives before returning control to the original policy~\cite{li2026spotter}.
RoboFAC appends a recovery subtask or an end-effector direction to the prompt, while the original VLA still produces the actions~\cite{ye2025robofac}.
Recova learns a recovery policy from failed scenes and human demonstrations, and likewise returns control once the scene becomes workable again~\cite{liu2026recova}.
In all of these methods, the original policy ultimately executes the remaining steps.
Embodied-R1.5 instead plans and corrects within a single model, without separating the decision model from the executor~\cite{yuan2026embodiedr}.
RoboDuo plans the remaining steps as atomic instructions and executes them with an atomic VLA trained only on short segments of successful demonstrations.
Both VLAs stay frozen at test time, and control does not return to the full-task VLA.
\section{Method}
\label{sec:method}

RoboDuo turns a single VLA rollout into a verified attempt.
A coding agent writes a program that gives a frozen full-task VLA one attempt on the original instruction and then tests the resulting observation with a completion check.
If the test fails, the program executes the remaining subtasks with an atomic VLA, which is trained on short clips that each contain a single complete skill.
The two VLAs produce every action, whereas the coding agent and the completion check only decide which VLA to query and which instruction to give it.
Sec.~\ref{sec:problem} formalizes the setting, and Sec.~\ref{sec:roboduo} presents the dual-system execution.
Sec.~\ref{sec:atomic} describes the segmentation of demonstrations into effect-complete clips, and Sec.~\ref{sec:components} describes how the atomic VLA, the coding agent, and the completion check are trained.
Fig.~\ref{fig:method} shows one execution.

\subsection{Problem Formulation}
\label{sec:problem}

A robot demonstration is a tuple
\begin{equation}
D = \left(\ell,\, o_{0:T},\, a_{0:T-1}\right),
\label{eq:demo}
\end{equation}
where $\ell$ is a language instruction, $o_t$ is the observation at time $t$, which includes camera images and the robot state, and $a_t$ is the action at that step.
A VLA predicts the next action from the current observation and the instruction,
\begin{equation}
\tilde{a}_t \sim \pi\!\left(a_t \mid o_t,\, \ell\right).
\label{eq:vla}
\end{equation}
Executing Eq.~\eqref{eq:vla} from $o_0$ yields a trajectory under a fixed instruction $\ell$.
At each step the policy sees a new observation $o_t$, but $\ell$ does not change and no check asks whether $o_t$ satisfies the intended goal.
The program in Sec.~\ref{sec:roboduo} supplies this test, while Eq.~\eqref{eq:vla} and its parameters remain unchanged.

\subsection{Dual-System Execution via Code}
\label{sec:roboduo}

RoboDuo separates a slow system that decides from a fast system that acts.
The fast system consists of two policies of the form in Eq.~\eqref{eq:vla}.
The full-task VLA $\pi_{\mathrm{full}}$ is conditioned on $\ell$, whereas the atomic VLA $\pi_{\mathrm{atom}}$ is conditioned on a single atomic instruction.
The slow system consists of the coding agent $\mathcal{C}$ and the completion check $J$.
At the start of an episode, $\mathcal{C}$ reads $\ell$ and $o_0$ and writes the program together with its atomic plan,
\begin{equation}
\left(P,\ \{(\ell_k, g_k)\}_{k=1}^{K}\right) = \mathcal{C}(\ell, o_0),
\label{eq:agent}
\end{equation}
where $P$ is the program, $\ell_k$ is the $k$-th atomic instruction, $g_k$ is its sub-goal in language, and $K$ is the number of atomic steps.
For example, $\ell_k$ may be \texttt{move\_to\_and\_grasp black\_bowl}, with the sub-goal $g_k$ ``the black\_bowl is grasped''.
The completion check, implemented as \texttt{is\_task\_complete}, predicts whether a goal holds in an observation,
\begin{equation}
J(o, g) \in \{0, 1\},
\qquad
g \in \{\ell\} \cup \{g_k\}_{k=1}^{K},
\label{eq:judge}
\end{equation}
where $J(o, g) = 1$ indicates that $g$ is satisfied in $o$.

By composing the two systems, RoboDuo selects each action as
\begin{equation}
\tilde{a}_t \sim
\begin{cases}
\pi_{\mathrm{full}}\!\left(a_t \mid o_t,\, \ell\right), & t < T_{\mathrm{full}},\\[3pt]
\pi_{\mathrm{atom}}\!\left(a_t \mid o_t,\, \ell_{k_t}\right), & t \ge T_{\mathrm{full}} \ \text{and}\ J\!\left(o_{T_{\mathrm{full}}}, \ell\right) = 0,
\end{cases}
\label{eq:roboduo_core}
\end{equation}
where $T_{\mathrm{full}}$ is the time at which the full-task attempt ends, and $k_t$ is the atomic step active at time $t$.
If $J(o_{T_{\mathrm{full}}}, \ell) = 1$, the episode terminates at $o_{T_{\mathrm{full}}}$.
Eq.~\eqref{eq:roboduo_core} retains the form of Eq.~\eqref{eq:vla} and changes only which policy is queried and under which instruction.

The program in Fig.~\ref{fig:method} implements this selection mechanism in two layers.
Layer~1 calls \texttt{vla\_execute\_fulltask}, which rolls out $\pi_{\mathrm{full}}$ from $o_0$ until $T_{\mathrm{full}}$, and then evaluates $J(o_{T_{\mathrm{full}}}, \ell)$.
If the check returns 1, the program stops without executing the atomic plan.
Otherwise, Layer~2 iterates over $\{(\ell_k, g_k)\}_{k=1}^{K}$, starting from $o_{T_{\mathrm{full}}}$.
For each step, it calls \texttt{vla\_execute\_atomic} on $\ell_k$ as long as $J(o, g_k) = 0$ and fewer than $N$ attempts have been made, with $N = 3$ in all experiments.
Each step starts from the observation left by the previous one, and $\pi_{\mathrm{full}}$ is never queried again.

The two systems communicate through code rather than through a latent vector or a language token.
A conditioning signal can change what a policy attends to, but it cannot end a trajectory, test its outcome, retry a step, or hand control to another policy.
In a program, these operations correspond to a function call, a conditional branch, and a bounded loop.
Moreover, $\mathcal{C}$ is called once per episode, and $J$ is called only after an attempt ends.
Between these calls, both VLAs run at their own control rates.

\begin{table}[t]
\centering
\caption{Primitive vocabulary of LIBERO-Atomic.
Every primitive takes a target \texttt{<object>}, and \texttt{place\_and\_release} also takes a relation (on, in, or under) and a \texttt{<target>}.
Pull and press are executed with the gripper open.
Calls count \texttt{vla\_execute\_atomic} invocations in the LIBERO-PRO evaluation (4{,}000 episodes, including retries).}
\label{tab:atomic-primitives}
\small
\setlength{\tabcolsep}{4pt}
\renewcommand{\arraystretch}{1.12}
\begin{tabular}{@{}c l p{0.36\linewidth} r@{}}
\toprule
\# & Primitive & Role & Calls \\
\midrule
1 & \texttt{move\_to\_and\_grasp}  & Approach and grasp an object        & 11{,}643 \\
2 & \texttt{place\_and\_release}   & Place a held object, then release   & 8{,}403 \\
3 & \texttt{move\_to\_and\_pull}   & Pull a drawer or door open          & 1{,}052 \\
4 & \texttt{rotate\_and\_release}  & Rotate a held knob, then release    & 488 \\
5 & \texttt{move\_to\_and\_press}  & Push a drawer or door shut, or press a button & 185 \\
\bottomrule
\end{tabular}
\end{table}

\begin{figure*}[t]
    \centering
    \includegraphics[width=1\linewidth]{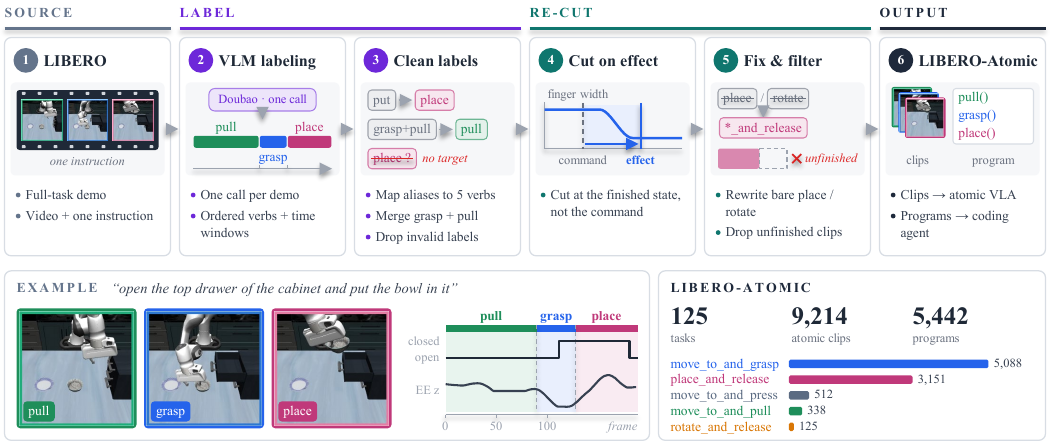}
    \caption{\textbf{Construction of LIBERO-Atomic.}
    An industry VLM labels each LIBERO demonstration with an ordered list of atomic verbs and time windows.
    Rule-based post-processing cleans these labels, and each clip is re-cut to end where its effect is complete.
    Each demonstration yields atomic clips with instructions for the atomic VLA, and a Python program of the same calls, which is wrapped into LIBERO-Code for the coding agent.}
    \label{fig:libero_atomic}
\end{figure*}

\subsection{Atomic Action Segmentation}
\label{sec:atomic}

RoboDuo recovers through short instructions that each name a single skill, so the training clips of $\pi_{\mathrm{atom}}$ must end when the effect of that skill is complete.
A full-task demonstration, however, is stored under one instruction and does not mark these boundaries.
We therefore decompose each demonstration into effect-complete clips,
\begin{equation}
D \;\mapsto\; \left\{\left(\ell_k,\; o_{s_k:e_k},\; a_{s_k:e_k-1}\right)\right\}_{k=1}^{K},
\label{eq:segment}
\end{equation}
where $s_k$ is the start of step $k$ and $e_k$ is the time at which its effect is complete.
The same boundaries later define the sub-goals written by $\mathcal{C}$ and the frame labels of $J$.

We instantiate this segmentation on LIBERO~\cite{liu2023libero} with the pipeline in Fig.~\ref{fig:libero_atomic}.
An industry VLM (Doubao Seed 2.1 Pro~\cite{bytedance2026seed21}) reads each video once and returns an ordered list of atomic verbs with time windows.
We write rules to map verb aliases onto the five primitives in Table~\ref{tab:atomic-primitives} and discard labels with an unknown verb or a missing target.
The rules also merge a grasp action followed by a pull on the same object into a single pull.
Since a labeled window usually closes when the command is issued, we move its end to $e_k$.
A grasp ends when the fingers stop closing, and a place or a rotate ends once the fingers have opened 20\,mm beyond the held width.
The end of a press or a pull is determined by an end-effector heading proxy.
Place and rotate segments are stored in their \texttt{*\_and\_release} forms, and clips whose effect never completes are removed.
The resulting dataset, LIBERO-Atomic, covers 125 tasks, with 9{,}214 clips and 5{,}442 programs.

\subsection{Training the Atomic VLA, Agent, and Judge}
\label{sec:components}

The clips and programs from Sec.~\ref{sec:atomic} train the atomic VLA $\pi_{\mathrm{atom}}$, the coding agent $\mathcal{C}$, and the completion check $J$.
$\pi_{\mathrm{full}}$ stays a released X-VLA or $\pi_{0.5}$ checkpoint, with its tokenizer, action expert, and training mixture unchanged.
The same program wraps either backbone by swapping $\pi_{\mathrm{full}}$ and the matching $\pi_{\mathrm{atom}}$.

\textbf{Atomic VLA.}
$\pi_{\mathrm{full}}$ is trained only on full-task instructions and therefore cannot follow the atomic instructions issued in Layer~2.
We thus train $\pi_{\mathrm{atom}}$ on the clips in Eq.~\eqref{eq:segment}, using the same architecture family as the chosen backbone, and freeze it at test time.
Table~\ref{tab:atomic-primitives} lists its five primitives, which account for 21{,}771 calls in the LIBERO-PRO evaluation.

\textbf{Coding agent.}
Each demonstration also yields the sequence of atomic calls that produced its clips.
We place this sequence behind the full-task test $J(o_{T_{\mathrm{full}}}, \ell)$ to form LIBERO-Code.
The target program $P_D$ thus first attempts $\ell$ and falls back to the atomic plan only if the check fails.
We fine-tune $\mathcal{C}$ by maximizing the likelihood of this target,
\begin{equation}
\mathcal{L}_{\mathcal{C}} = -\,\mathbb{E}_{D}\left[\log p_{\mathcal{C}}\!\left(P_D \mid \ell,\, o_0\right)\right].
\label{eq:loss_agent}
\end{equation}
Because every target places the atomic plan behind the full-task test, $\mathcal{C}$ learns to write a full-task attempt first rather than to decompose the task from the first step.
The atomic calls share the verb vocabulary of the clips, so the instructions written by $\mathcal{C}$ remain within the training distribution of $\pi_{\mathrm{atom}}$.


\paragraph{Completion check.}
LIBERO-Judge converts the same boundaries into frame labels.
For a frame $o_t$ in the window of step $k$,
\begin{equation}
y_{t,k}=
\begin{cases}
1, & \text{if } t\ge e_k,\\
0, & \text{otherwise.}
\end{cases}
\end{equation}
Each triple $(o_t,g_k,y_{t,k})$ forms one question--answer pair.
$J$ is a vision-language classifier trained on these pairs to answer whether $g_k$ holds in $o_t$.
A step is therefore accepted under the same criterion that ended the corresponding training clip of $\pi_{\mathrm{atom}}$.
\section{Experiments}
\label{sec:experiments}

We organize the experiments around four research questions:
\begin{itemize}
    \item \textbf{RQ1:} Does RoboDuo improve frozen full-task VLAs on standard manipulation tasks?
    \item \textbf{RQ2:} Does RoboDuo improve robustness under position and task perturbations?
    \item \textbf{RQ3:} How does the choice of coding agent affect recovery?
    \item \textbf{RQ4:} Does RoboDuo transfer to a physical robot?
\end{itemize}
Unless noted otherwise, we report closed-loop task success rate (\%).

\subsection{Overall Performance on LIBERO}
\label{sec:libero}

\paragraph{Experimental setup.}
We evaluate on  four LIBERO suites~\cite{liu2023libero}, Spatial, Object, Goal, and Long, with 10 tasks per suite and 50 trials per task.
RoboDuo wraps two frozen full-task backbones, X-VLA~\cite{zheng2025xvla} and $\pi_{0.5}$~\cite{intelligence2025pi05}, each trained on the four evaluation suites or on five suites that add LIBERO-90.
We test every checkpoint alone and under RoboDuo with identical full-task weights.
A Qwen3.5-2B~\cite{yang2025qwen3} fine-tuned on LIBERO-Code and LIBERO-Judge (Sec.~\ref{sec:components}) serves as both the coding agent $\mathcal{C}$ and the completion check $J$.
Each backbone has a matching atomic VLA trained on LIBERO-Atomic (100k steps for X-VLA, 30k for $\pi_{0.5}$) on 8 NVIDIA A800 GPUs.
At inference, the program makes one full-task attempt and, if $J$ reports failure, runs the atomic plan with at most three attempts per step.
Table~\ref{tab:libero} also lists published LIBERO results of recent VLAs for context.

\paragraph{Results.}
Table~\ref{tab:libero} shows that RoboDuo improves both frozen backbones without lowering any suite, with gains that track how much unfinished work the base leaves behind.

Under four-suite training, Long drives most of the improvement.
X-VLA rises from 95.25 to 96.40 on average ($87.4\rightarrow 91.6$ on Long; Spatial $+0.4$; Object and Goal unchanged), and $\pi_{0.5}$ from 97.25 to 98.40 ($93.6\rightarrow 96.2$ on Long, with smaller gains elsewhere).
On Long, RoboDuo removes 33\% of the failures of X-VLA and 41\% of those of $\pi_{0.5}$.
Longer task chains amplify cascade failures after a single bad step, so recovery helps most there.
RoboDuo\,+\,$\pi_{0.5}$ reaches 98.40, slightly above the five-suite $\pi_{0.5}$ base (98.30), without updating any weights.

Under five-suite training, both bases are already stronger on Long, so residual gains shrink: X-VLA $97.65\rightarrow 97.95$ and $\pi_{0.5}$ $98.30\rightarrow 98.45$, again mainly from Long.
These results show that a program that attempts the full task and falls back to the atomic VLA when $J$ reports failure raises success on both frozen backbones, answering RQ1.

\begin{table}[t]
\centering
\caption{LIBERO success rate (\%).
\emph{Upper block:} published numbers from the cited papers (not rerun).
\emph{Middle / lower blocks:} our evaluation under four- and five-suite training, each backbone alone and with RoboDuo on the same weights.
Bold marks the best average within each of our training blocks.}
\label{tab:libero}
\footnotesize
\setlength{\tabcolsep}{3.5pt}
\begin{tabular}{lccccc}
\toprule
Method & Spatial & Object & Goal & Long & Avg. \\
\midrule
$\pi_0$~\cite{black2024pi0} & 98.0 & 96.8 & 94.4 & 88.4 & 94.4 \\
$\pi_{0.5}$~\cite{intelligence2025pi05} & 98.8 & 98.2 & 98.0 & 92.4 & 96.9 \\
InternVLA-M1~\cite{internvla2025m1} & 98.0 & 99.0 & 93.8 & 92.6 & 95.9 \\
GR00T-N1.7~\cite{bjorck2025gr00t} & 97.7 & 98.5 & 97.5 & 94.4 & 97.0 \\
OpenVLA-OFT~\cite{kim2025oft} & 97.6 & 98.4 & 97.9 & 94.5 & 97.1 \\
Fast-WAM~\cite{yuan2026fastwam} & 98.2 & 100.0 & 97.0 & 95.2 & 97.6 \\
Motus~\cite{bi2025motus} & 96.8 & 99.8 & 96.6 & 97.6 & 97.7 \\
Being-H0.5~\cite{luo2026beingh05} & 99.2 & 98.2 & 99.0 & 96.2 & 98.2 \\
\midrule
\multicolumn{6}{l}{\emph{Four-suite training (our evaluation)}} \\
X-VLA (reproduced) & 96.2 & 99.6 & 97.8 & 87.4 & 95.25 \\
RoboDuo\,+\,X-VLA & 96.6 & 99.6 & 97.8 & 91.6 & 96.40 \\
$\pi_{0.5}$ (reproduced) & 99.4 & 98.0 & 98.0 & 93.6 & 97.25 \\
RoboDuo\,+\,$\pi_{0.5}$ & 99.6 & 99.0 & 98.8 & 96.2 & \textbf{98.40} \\
\midrule
\multicolumn{6}{l}{\emph{Five-suite training (our evaluation)}} \\
X-VLA (reproduced) & 97.2 & 99.0 & 98.6 & 95.8 & 97.65 \\
RoboDuo\,+\,X-VLA & 97.4 & 99.2 & 98.6 & 96.6 & 97.95 \\
$\pi_{0.5}$ (reproduced) & 99.4 & 99.8 & 98.6 & 95.4 & 98.30 \\
RoboDuo\,+\,$\pi_{0.5}$ & 99.4 & 99.8 & 98.6 & 96.0 & \textbf{98.45} \\
\bottomrule
\end{tabular}
\end{table}

\begin{table*}[t]
\centering
\caption{LIBERO-PRO success rate (\%). S is the position perturbation (\texttt{swap}) and T is the task perturbation.
\emph{Upper block:} published numbers under the official file-name prompting protocol.
Mean is the arithmetic mean of the eight columns. Bold marks the best result in each column.}
\label{tab:liberopro}
\scriptsize
\setlength{\tabcolsep}{2.8pt}
\begin{tabular}{lccccccccc}
\toprule
Method & Spatial-S & Spatial-T & Object-S & Object-T & Goal-S & Goal-T & Long-S & Long-T & Mean \\
\midrule
OpenVLA & 0.0 & 0.0 & 0.0 & 0.0 & 0.0 & 0.0 & 0.0 & 0.0 & 0.00 \\
$\pi_0$ & 0.0 & 0.0 & 0.0 & 0.0 & 0.0 & 0.0 & 0.0 & 0.0 & 0.00 \\
$\pi_{0.5}$ & 20.0 & 1.0 & 17.0 & 1.0 & 38.0 & 0.0 & 8.0 & 1.0 & 10.75 \\
MolmoAct & 0.0 & 0.0 & 6.0 & 0.0 & 0.0 & 0.0 & 0.0 & 6.0 & 1.50 \\
NORA & 0.0 & 0.0 & 0.0 & 0.0 & 0.0 & 0.0 & 0.0 & 0.0 & 0.00 \\
X-VLA & 0.0 & 0.0 & 2.0 & 8.0 & 1.0 & 9.0 & 0.0 & 10.0 & 3.75 \\
\midrule
\multicolumn{10}{l}{\emph{Four-suite training (our evaluation)}} \\
X-VLA (reproduced) & 0.0 & 18.0 & 0.0 & 0.0 & 2.6 & 9.2 & 0.0 & 9.8 & 4.95 \\
RoboDuo\,+\,X-VLA & 0.2 & 18.0 & 0.0 & 0.0 & 3.2 & 10.2 & 0.4 & 13.6 & 5.70 \\
$\pi_{0.5}$ (reproduced) & 63.6 & 60.6 & 20.0 & 25.2 & 39.0 & 24.2 & 12.6 & 11.2 & 32.05 \\
RoboDuo\,+\,$\pi_{0.5}$ & 67.8 & \textbf{61.0} & 31.0 & \textbf{55.8} & 39.8 & 26.2 & \textbf{15.6} & 17.4 & 39.33 \\
\midrule
\multicolumn{10}{l}{\emph{Five-suite training (our evaluation)}} \\
X-VLA (reproduced) & 2.0 & 16.0 & 2.0 & 6.0 & 0.0 & 12.0 & 0.0 & 20.0 & 7.25 \\
RoboDuo\,+\,X-VLA & 2.0 & 18.0 & 4.0 & 6.0 & 0.0 & 12.0 & 0.0 & 26.0 & 8.50 \\
$\pi_{0.5}$ (reproduced) & 70.0 & 55.0 & 49.2 & 32.2 & 44.2 & 34.6 & 13.6 & 49.8 & 43.58 \\
RoboDuo\,+\,$\pi_{0.5}$ & \textbf{71.6} & 55.0 & \textbf{50.8} & 42.2 & \textbf{45.4} & \textbf{35.0} & 15.2 & \textbf{50.2} & \textbf{45.68} \\
\bottomrule
\end{tabular}
\end{table*}

\begin{table*}[t]
\centering
\caption{Effect of the coding agent on LIBERO-PRO (\%). Every RoboDuo row uses frozen $\pi_{0.5}$ as the full-task VLA, with the same atomic VLA and retry budget.
GPT denotes GPT-6-Astra.
Mean-6 averages Spatial, Object, and Goal; Mean-8 averages all eight columns.
Cap-X and Pigey are published references that omit Long.
GPT cells use 50 episodes per cell.
Bold marks the best result within each backbone block (Cap-X and Pigey excluded).}
\label{tab:exp_agent}
\scriptsize
\setlength{\tabcolsep}{2.6pt}
\begin{tabular}{lcccccccccc}
\toprule
Coding agent & Spatial-S & Spatial-T & Object-S & Object-T & Goal-S & Goal-T & Long-S & Long-T & Mean-6 & Mean-8 \\
\midrule
Cap-X~\cite{fu2025capx}  & 12.0 & 14.0 & 22.0 & 18.0 & 26.0 & 17.0 & -- & -- & 18.17 & -- \\
Pigey~\cite{galanti2026pigey}  & 66.0 & 80.0 & 54.0 & 54.0 & 44.0 & 22.0 & -- & -- & 53.33 & -- \\
\midrule
\multicolumn{11}{l}{\emph{Four-suite training (our evaluation)}} \\
None (frozen $\pi_{0.5}$) & 63.6 & 60.6 & 20.0 & 25.2 & 39.0 & 24.2 & 12.6 & 11.2 & 38.77 & 32.05 \\
LIBERO-Code fine-tuned    & 67.8 & 61.0 & 31.0 & \textbf{55.8} & 39.8 & 26.2 & 15.6 & 17.4 & 46.93 & 39.33 \\
GPT                       & 68.0 & 60.0 & \textbf{36.0} & 54.0 & 38.0 & 34.0 & \textbf{20.0} & 22.0 & 48.33 & 41.50 \\
GPT\,+\,TAMP              & \textbf{74.0} & \textbf{64.0} & 30.0 & 54.0 & \textbf{40.0} & \textbf{36.0} & \textbf{20.0} & \textbf{26.0} & \textbf{49.67} & \textbf{43.00} \\
\midrule
\multicolumn{11}{l}{\emph{Five-suite training (our evaluation)}} \\
None (frozen $\pi_{0.5}$) & 70.0 & 55.0 & 49.2 & 32.2 & 44.2 & 34.6 & 13.6 & 49.8 & 47.53 & 43.58 \\
LIBERO-Code fine-tuned    & 71.6 & 55.0 & 50.8 & 42.2 & 45.4 & 35.0 & 15.2 & 50.2 & 50.00 & 45.68 \\
GPT                       & 72.0 & 56.0 & \textbf{60.0} & \textbf{54.0} & 48.0 & 38.0 & 16.0 & \textbf{54.0} & 54.67 & 49.75 \\
GPT\,+\,TAMP              & \textbf{76.0} & \textbf{58.0} & \textbf{60.0} & 48.0 & \textbf{58.0} & \textbf{42.0} & \textbf{20.0} & 50.0 & \textbf{57.00} & \textbf{51.50} \\
\bottomrule
\end{tabular}
\end{table*}

\subsection{Robustness on LIBERO-PRO}
\label{sec:liberopro}

\paragraph{Experimental setup.}
We evaluate the same checkpoints zero-shot on LIBERO-PRO~\cite{zhou2025liberopro}.
For each of the four suites, a position perturbation (S, \texttt{swap}) and a task perturbation (T) yield eight conditions.
Each of our cells aggregates 10 tasks $\times$ 50 trials (500 episodes, seed 42), for four- and five-suite backbones alone and under RoboDuo.
All our rows condition the policy on the BDDL \texttt{:language} field, which matches the scored goal predicate; the official leaderboard uses the file-name instruction instead.
Under file-name prompting, the policy is asked for the original task but scored on the perturbed one, so task-perturbation success is near zero by construction.

\paragraph{Results.}
Table~\ref{tab:liberopro} shows that RoboDuo improves robustness when the frozen backbone already succeeds on some perturbed episodes, and no cell falls below its base.

Under four-suite training, RoboDuo raises the eight-condition mean of $\pi_{0.5}$ from 32.05 to 39.33.
The largest gains are on Object-T ($+30.6$) and Object-S ($+11.0$), where recovery can build on a partly completed attempt.
Long-T and Spatial-S improve more modestly, and Spatial-T stays within noise.
X-VLA moves only from 4.95 to 5.70: conditions where its base is zero stay at or below $0.4$, and the largest change is Long-T ($9.8\rightarrow 13.6$).

Under five-suite training, the $\pi_{0.5}$ base is already stronger (43.58), and RoboDuo still adds $2.10$ points to reach 45.68, again led by Object-T ($32.2\rightarrow 42.2$).
X-VLA remains weak ($7.25\rightarrow 8.50$), with most of the gain on Long-T ($20.0\rightarrow 26.0$).
Because RoboDuo reuses frozen policies, it amplifies skills the backbone partly has rather than creating new ones.
These results answer RQ2.

\subsection{Effect of the Coding Agent}
\label{sec:exp_agent}

\paragraph{Experimental setup.}
To answer RQ3, we fix the frozen $\pi_{0.5}$ full-task and atomic policies and the retry budget, and swap only the slow-system pair $(\mathcal{C},\,J)$.
We evaluate on the eight LIBERO-PRO conditions of Sec.~\ref{sec:liberopro}, with $\pi_{0.5}$ trained on four or five suites.
We compare two agents that each write the program and act as $J$: the fine-tuned Qwen3.5-2B, and GPT-6-Astra~\cite{galbot2026astra} with few-shot code prompts and the current camera views.
We also run GPT-6-Astra in a geometric TAMP scaffold that may append OSC primitives after a failed full-task VLA call, with GPT-6-Astra again acting as $J$.
Unlike Pigey~\cite{galanti2026pigey}, which selects tools step by step and observes the simulator \texttt{task\_done} flag, our TAMP agent generates code, grounds objects with SAM3, and relies on $J$ rather than this privileged signal.

\paragraph{Results.}
Table~\ref{tab:exp_agent} shows that coding agents are not interchangeable: with the same frozen VLAs, the eight-condition mean varies by up to $5.8$ points across agents.

Every agent raises the mean above the frozen backbone.
Under four-suite training, the fine-tuned agent lifts the mean from $32.05$ to $39.33$, GPT reaches $41.50$, and GPT\,+\,TAMP reaches $43.00$.
Relative to the fine-tuned agent, GPT gains mainly on Goal-T, Object-S, and both Long conditions.
Relative to GPT alone, TAMP gains most on Spatial-S ($+6.0$) and on Spatial-T and Long-T ($+4.0$ each), while Object-S drops by $6.0$.

Under five-suite training, the same ordering holds ($43.58\rightarrow 45.68\rightarrow 49.75\rightarrow 51.50$).
Here TAMP mainly helps position perturbations (S mean $+4.5$ over GPT, led by Goal-S at $+10.0$; T mean $-1.0$).
Because GPT cells use 50 episodes, gaps of a few points are trends rather than strict rankings.
The VLAs are identical across rows, so these differences come from the generated programs, which answers RQ3.

\begin{figure*}[t]
\centering
\includegraphics[width=\textwidth]{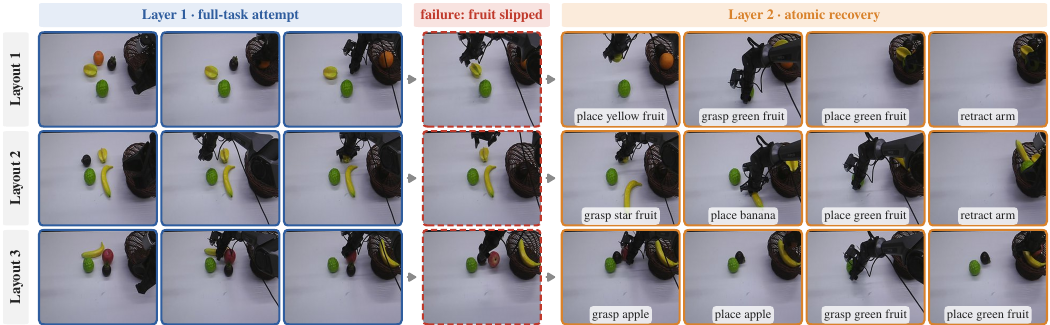}
\caption{Real-world RoboDuo on a bimanual AgileX Piper for fruit arrangement across three layouts.
Blue: frozen full-task VLA; red: the completion check fails and control switches; orange: recovery by the frozen atomic VLA.}
\label{fig:real_case}
\end{figure*}



\subsection{Real-World Experiments}
\label{sec:real}

\paragraph{Experimental setup.}
We deploy RoboDuo on a bimanual AgileX Piper for fruit arrangement across three layouts (Fig.~\ref{fig:real_case}).
A coding agent writes the program once at the start of each episode, and the frozen full-task VLA attempts the instruction first.
A gripper monitor and a scene check then decide whether to hand the remaining steps to the frozen atomic VLA.
GPT-6-Astra judges grasps and basket counts from the overhead and wrist views.
The atomic VLA is invoked after a reported drop, a hold longer than six chunks, or a placement that does not increase the basket count.
Both VLAs are fine-tuned from the same pretrained X-VLA checkpoint, on teleoperated demonstrations and on clips segmented by Sec.~\ref{sec:atomic}, respectively, and stay frozen at test time.
Appendix gives the control stack, data statistics, and full switching rules.

\paragraph{Results.}
Table~\ref{tab:real} shows that the same test-time controller transfers to the robot.
On layout~1, RoboDuo raises success from $40\%$ to $60\%$ and mean fruits in the basket from $2.4$ to $3.0$ over $10$ trials.
On layout~2, success rises from $30\%$ to $60\%$ and fruits from $1.9$ to $2.8$.
On layout~3, success stays at $40\%$, but mean fruits rise from $1.8$ to $2.6$.
Pooled over all layouts, success increases from 11/30 to 16/30 trials.
In most RoboDuo trials, the atomic VLA is called after a place timeout, a detected drop, or a scene check reporting no progress, and then finishes the remaining grasp--place steps.
With $10$ trials per layout these differences are indicative rather than conclusive, but the gains require no on-robot update to either VLA, which answers RQ4.

\begin{table}[t]
\centering
\caption{Real-world fruit arrangement on a bimanual AgileX Piper.
Success rate and mean fruit count over 10 trials per method and layout.
Success requires all four fruits in the basket.}
\label{tab:real}
\footnotesize
\setlength{\tabcolsep}{6pt}
\begin{tabular}{clcc}
\toprule
Layout & Method & Success (\%) & Fruits (/4) \\
\midrule
1 & Full-task VLA & 40 & 2.4 \\
  & RoboDuo       & 60 & 3.0 \\
\midrule
2 & Full-task VLA & 30 & 1.9 \\
  & RoboDuo       & 60 & 2.8 \\
\midrule
3 & Full-task VLA & 40 & 1.8 \\
  & RoboDuo       & 40 & 2.6 \\
\bottomrule
\end{tabular}
\end{table}

\section{Conclusion}
\label{sec:conclusion}

We presented RoboDuo, a test-time controller that uses code as the interface between two frozen VLAs.
A coding agent writes a program that gives the full-task VLA one attempt and checks the outcome with a completion check $J$.
On failure, the program hands the remaining steps to an atomic VLA, and control does not return to the full-task policy.
The same program wraps both X-VLA and $\pi_{0.5}$ without changing their weights.

Across benchmarks, the gains follow how much recoverable work the base policy leaves behind.
On LIBERO, RoboDuo improves both backbones, especially on long-horizon tasks, without lowering performance on any suite.
On LIBERO-PRO, it remains effective under position and task perturbations without retraining either VLA.
With the VLAs fixed, a stronger coding agent yields further gains, suggesting that RoboDuo can improve with better coding agents.
On a bimanual robot, the same controller improves both task success and the number of fruits placed, without on-robot updates.
These results suggest that a failed episode is better finished as a sequence of atomic skills than continued under the original instruction.

As both VLAs stay frozen, RoboDuo cannot recover skills that neither policy has, and its gains are bounded by the accuracy of the completion check.
Future work may include stronger error-recovery capabilities and broader real-robot validation.

\clearpage
{
    \small
    \bibliographystyle{ieeenat_fullname}
    \bibliography{main}
}


\end{document}